\documentclass{article}
\usepackage[preprint]{corl_2026}
\makeatletter
\providecommand{\@notice}{\begingroup\renewcommand{\thefootnote}{}\footnotetext{\@noticestring}\endgroup}
\makeatother
\usepackage{graphicx}
\usepackage{booktabs}
\usepackage{tabularx}
\usepackage{multirow}
\usepackage{amsmath,amssymb}
\usepackage{placeins}
\usepackage{tikz}
\usepackage{fontawesome5}

\usetikzlibrary{positioning,arrows.meta}
\title{Grip, Trace, Write: Force-Aware\\Imitation and Reinforcement Learning\\for High-DoF Robotic Handwriting}
\author{Rongqian Chen \And Yiding Chen \And Jijie Zhou \And Tom Z. Jiahao \AND
\textmd{Daxo Robotics}\\
\textmd{\raisebox{-0.1em}{\faGithub}\ \texttt{\href{https://dexwriting.github.io/}{https://dexwriting.github.io/}}}}

\begin{document}
\maketitle
\vspace{-3mm}
\begin{abstract}
Most robotic handwriting rigidly mounts the pen and reduces writing to trajectory tracking. A freely held pen instead couples grasp stability, tip motion, and contact force.
We present \emph{Grip, Trace, Write}, a force-aware simulation environment and learning framework for high-DoF handwriting. A two-branch actor separates pen retention from arm--wrist writing while preserving their closed-loop coupling. Training learns a grasp controller, initializes writing from inverse-kinematics demonstrations by behavior cloning, and improves writing through reinforcement learning with rewards on ink, force, time, and retention. Across diverse tasks, the policy improves completion, precision, force regulation, and efficiency over behavior cloning and inverse kinematics, and generalizes without fine-tuning to unseen English and Chinese content. These results establish force-aware handwriting as a testbed for dexterous tool use.
\end{abstract}

\section{Introduction}


Writing with a freely held pen is a compact testbed for dexterous tool use.
It requires a hand to stabilize a slender object through a precision grip while the arm and wrist transport its tip and regulate its contact with a surface.
Human manual dexterity similarly combines stable multi-finger grasping, fine motion, and force-sensitive interaction with tools~\cite{kivell2023form}.

This coupling makes handwriting fundamentally different from moving a rigidly mounted pen along a geometric path.
A desired tip trajectory alone cannot determine whether ink is deposited: insufficient force produces hovering or broken strokes, whereas excessive force can distort the trace, alter the pen pose, or destabilize the grasp.
With a freely held pen, finger--pen contact, pen--sheet contact, and arm motion therefore affect one another.
Existing systems commonly avoid this coupling by rigidly mounting the pen or brush to an arm~\cite{mueller2013robotic,wu2018towards,luo2024callirewrite,schaldenbrand2023frida}, or address only limited writing behaviors with multi-fingered hands~\cite{rombokas2011task,zhao2025tactile}.
They typically treat writing as trajectory tracking~\cite{stewart2026rapid}, with contact force uncontrolled or maintained around a fixed setpoint, and evaluate only a small set of strokes or glyphs~\cite{hu2025slender,zhao2025tactile}.
Consequently, sustained force-regulated writing with a freely held pen and a high-DoF hand remains unresolved.

We address this problem in a force-aware simulation environment for robotic handwriting.
A learned grasp MLP first establishes a stable hand--pen interface.
Given a requested 5D cursor, an inverse-kinematics (IK) controller generates nominal arm--wrist commands while the learned grasp holds the pen; these rollouts initialize a trajectory-conditioned writing policy through behavior cloning (BC).
BC provides broad coverage of the writing repertoire but only imitates nominal IK actions.
We therefore improve writing through RL with rewards defined on task outcomes: ordered ink coverage, off-target ink, force tracking, elapsed time, and pen retention.
This separation makes the roles explicit: grasp learning resolves tool stabilization, IK-guided BC makes high-DoF writing tractable, and RL corrects contact-induced errors that action imitation does not optimize.

Our contributions are: 
(1) a force-aware simulation environment for high-DoF robotic handwriting with a freely held pen, contact dynamics, and diverse tracing and writing tasks; 
(2) a two-branch controller that separates pen retention from arm--wrist writing while retaining their closed-loop physical coupling; and 
(3) a grasp-to-writing learning procedure that uses IK-guided BC for broad initialization and task-outcome RL to improve trace completion, ink precision, contact-force accuracy, and execution efficiency.

\paragraph{Related work.}
Prior robot-writing and drawing systems typically rigidly mount a pen or brush to an arm, reducing the problem to end-effector trajectory generation or ink modeling~\cite{mueller2013robotic,wu2018towards,luo2024callirewrite,schaldenbrand2023frida}. Writing with a pen held in a multi-fingered hand remains rare and limited to single strokes or simple shapes, using synergies~\cite{rombokas2011task}, online Jacobian estimation~\cite{stewart2026rapid}, or tactile RL with compliant wrist control~\cite{zhao2025tactile}. 
Learned in-hand policies reorient and spin objects without external contact~\cite{openai2020shadow,handa2023dextreme,wang2024penspin}, track a held stick’s tip without touching a surface~\cite{hu2025slender}, or bring a held tool into contact without regulating force~\cite{kedia2026simtoolreal}.
Initializing from demonstrations and refining with RL is effective for dexterous control~\cite{rajeswaran2018dapg,qin2022dexmv}; we apply it to the full writing repertoire with outcome rewards on ink, force, and pen retention.

\section{Method}
\subsection{Environment and control setup}
\begin{figure}[!h]
\centering
\includegraphics[width=\textwidth]{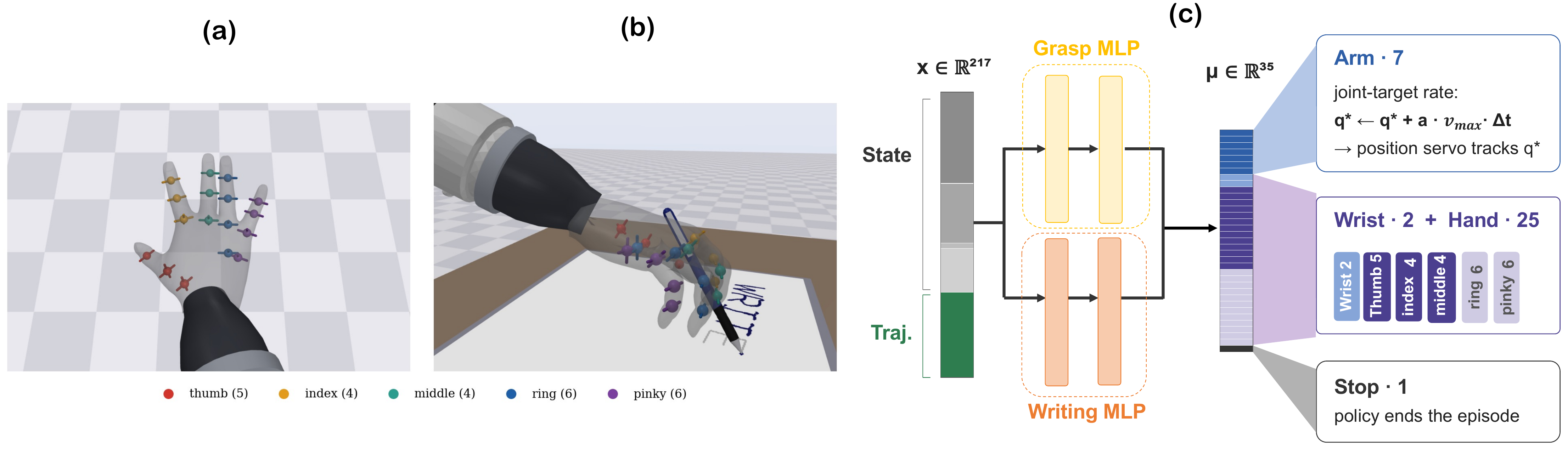}
\caption{\textbf{Hand kinematics and two-branch actor.} (a)(b) Spheres mark the finger joints, colored by finger, with rods along the
hinge axes; the skin is rendered translucent. The hand has 25
finger joints, in addition to 2 wrist and 7 arm joints. (c)
Actor model reads both the same state and trajectory. A fixed output mask selects finger actions from the grasp branch and UR5e arm, wrist, and stop actions from the writing branch. The branches interact through the held pen and live state.}
\label{fig:architecture}
\end{figure}

Our MuJoCo setup~\cite{todorov2012mujoco} mounts a high-DoF anthropomorphic hand (Fig.~\ref{fig:architecture}(a)(b)) on a UR5e and places a free pen in a tripod grasp without virtual assistance. The hand has 25 finger joints, in addition to 2 wrist and 7 arm joints. The suite combines three retention conditions (static, moving, and contact scribbling), three tracing families (line, circle, and varied shapes), and handwriting of digits, capital letters, English words, and Chinese characters. Training covers all 10 digits, 26 capital letters, 15 English words, and 118 Chinese characters with several fonts. Every episode resamples placement, scale, slant, rotation, and target-force profile. Zero-shot tests add 20 held-out English words, 10 four-character Chinese idioms, and 122 Chinese characters with at least seven strokes.  Appendix~\ref{app:task_config} gives complete task and interface definitions.
\vspace{-3mm}

\subsection{Two-branch policy}

The actor has separate grasp and writing MLPs (Fig.~\ref{fig:architecture}(c)). A fixed mask takes finger actions from the grasp trunk and arm, wrist, and stop actions from the writing trunk. This split follows the structure of the problem: pen retention can be learned from holding interactions before any glyph is specified, whereas tip transport can use IK once a stable grasp exists. Both trunks still read the same live state, so the mask separates supervision and action roles without breaking the physical coupling among hand, pen, and arm. The transferred grasp controller is folded into the new input representation; its preservation during final RL is described below.

\subsection{Training workflow}
\label{sec:training_workflow}
\begin{figure}[t]
\centering
\includegraphics[width=\textwidth]{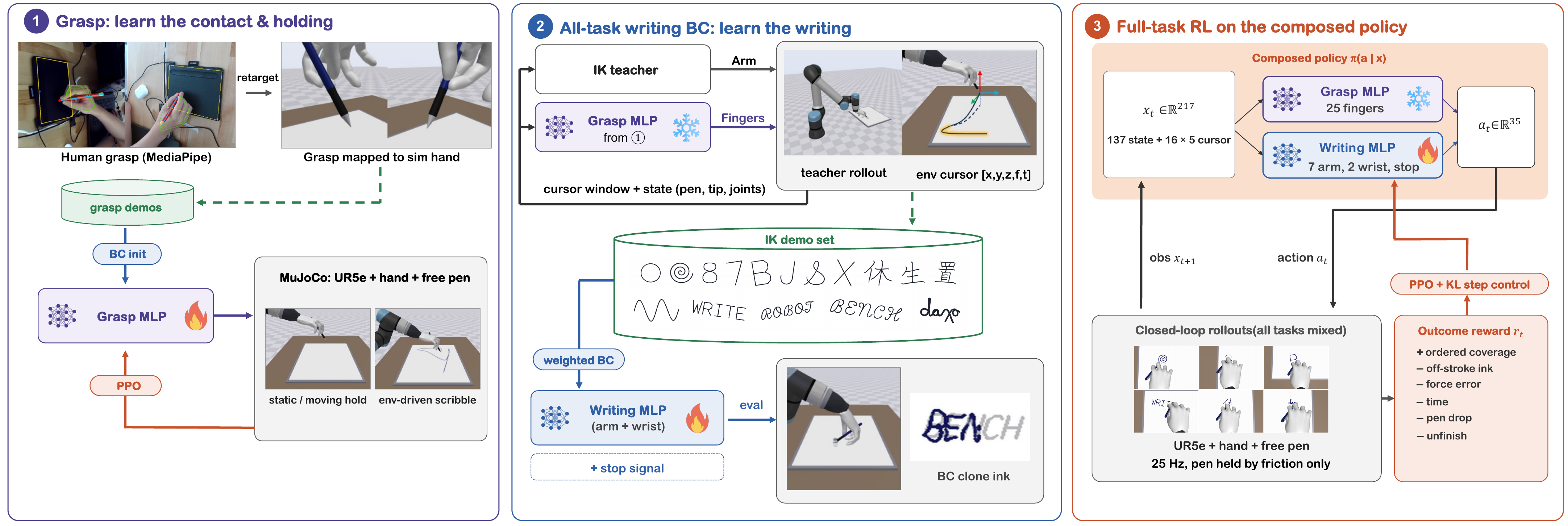}
\caption{\textbf{Training pipeline.} Panels 1--3 correspond to the three stages in Sec.~\ref{sec:training_workflow}: grasp BC and hold RL, all-task writing BC with the learned grasp, and full-task outcome RL on the composed policy.}
\label{fig:pipeline}
\end{figure}

\paragraph{Grasp BC and RL.}
Stable pen retention is a prerequisite for meaningful handwriting learning. When the pen slips, missing or misplaced ink can arise from either failed finger contact or incorrect arm motion, so the writing objective does not clearly identify which subsystem should change. We therefore learn the grasp branch before the writer. BC on tripod-hold demonstrations quickly initializes a valid multi-finger contact configuration; pen-hold RL then improves retention under static, arm-driven, and environment-driven motion, covering contact disturbances absent from the demonstrations. Transferring this controller gives subsequent IK data collection and writing optimization a consistent, physically stable hand--pen interface.

\paragraph{All-task writing BC.}
Once the grasp is reliable, the remaining problem is to map a requested trajectory to arm and wrist motion under that particular hand--pen contact. We therefore run IK while the transferred grasp controls the fingers. This produces demonstrations in the same compliant, freely held system that the learned writer will encounter, avoiding a mismatch with data collected using an idealized fixed pen. Clean rollouts teach nominal motion; perturbing the executed action moves the system into nearby off-nominal states while retaining the clean IK command as the recovery target~\cite{laskey2017dart}. Because traces, glyphs, words, and Chinese writing are all presented through the same trajectory interface, we pool them to learn a geometry-conditioned controller rather than separate content-specific policies. The writing branch minimizes
\begin{equation}
\mathcal L_{\mathrm{BC}}=\mathbb E_{(x,a^*)}\sum_{d=1}^{9}
w_d\bigl(\mu_d(x)-a_d^*/s_d\bigr)^2,\quad
w_d=\operatorname{clip}\!\left[\left(\frac{\operatorname{median}_j\sigma_j}
{\max(\sigma_d,10^{-3})}\right)^2,1,100\right],
\label{eq:bc_main}
\end{equation}
where $s_d$ scales action dimension $d$ and $\sigma_d$ is its label standard deviation. Inverse-variance weighting balances coordinates with different natural ranges, and fitting the stop output on the clone's own rollouts teaches termination under its induced state distribution. This stage gives the complete system a broad, stable writing initialization, but it still optimizes agreement with IK rather than the physical trace.

\paragraph{Full-task outcome RL.}
BC solves initialization, but leaves two gaps: matching IK commands does not directly optimize ink and force, and small execution errors can carry the clone beyond the demonstrated states. We therefore refine the composed controller with PPO on the complete task mixture. The grasp mean remains frozen so drawing reward cannot trade away the reliable contact strategy, and frozen observation statistics prevent normalization drift from changing the transferred policy. The writing mean and Gaussian exploration parameters remain trainable, while both branches stay closed-loop through their shared state and the physical pen. Let $C_t$ be ordered coverage within 1\,mm, $P_t$ normalized off-target ink, $E_t$ running force MAE, $\tau_t$ normalized elapsed time, $p_t$ the pen tip, and $q_t$ the first unfinished target point at pressing depth. The outcome reward is
\begin{equation}
r_t=\lambda_C\Delta C_t-\lambda_P\Delta P_t-\lambda_E\Delta E_t
-\lambda_\tau\Delta\tau_t
+\lambda_q\bigl(\|p_{t-1}-q_t\|-\|p_t-q_t\|\bigr)-D_t,
\label{eq:reward_main}
\end{equation}
where all $\lambda$ weights are positive and $D_t$ penalizes drops and premature termination. Each term targets a distinct failure mode: ordered coverage rewards forward stroke progress, off-target ink rejects indiscriminate scribbling, and the pressing-depth point $q_t$ prevents progress by hovering above the sheet. Force and time distinguish meaningful contact from mere touching and favor efficient completion, while $D_t$ preserves the pen-retention prerequisite established in the first stage. Crucially, there is no BC or IK-timing term, so IK provides an initialization rather than a performance ceiling. Appendix~\ref{app:implementation} gives the numerical weights, penalty construction, PPO schedule, and optimization details.
\vspace{-2mm}
\section{Experiments}
\vspace{-2mm}
\paragraph{Protocol.}
We compare IK, all-task BC, and final RL on eight primary tasks with 50 matched seeds each. Policies execute deterministic means and their own stop outputs. We report ordered completion and precision within 1\,mm, force MAE, time, shape distances, and pen drops; generalization tests use unseen English and Chinese content without fine-tuning.

\subsection{Writing performance}
\vspace{-2mm}
\begin{table}[!h]
\centering\scriptsize
\setlength{\tabcolsep}{1.5pt}
\renewcommand{\arraystretch}{0.88}
\begin{tabular*}{\textwidth}{@{\extracolsep{\fill}}clrrrrrrrr|rrr|r@{}}
\toprule
\multirow{2}{*}{Controller} & \multirow{2}{*}{  Metric} & \multicolumn{8}{c|}{Primary tasks} & \multicolumn{3}{c|}{Unseen} & \multirow{2}{*}{Mean} \\
& & Line & Circle & Shape & Digits & Letters & Word & Cursive & Chinese & \shortstack{word} & Idiom & \shortstack{Chinese} & \\
\midrule
\multirow{4}{*}{IK} & Completion (\%) $\uparrow$ & \textbf{98.2} & 98.8 & 94.9 & 90.6 & 92.7 & 85.4 & 83.4 & 91.7 & 87.7 & 92.1 & 87.7 & 92.0 \\
 & Precision (\%) $\uparrow$ & 95.7 & 96.4 & 82.6 & 75.8 & 82.2 & 71.5 & 61.6 & 78.7 & 71.1 & 82.1 & 71.7 & 80.6 \\
 & Force MAE (N) $\downarrow$ & 0.558 & 0.537 & 0.529 & 0.531 & 0.533 & 0.582 & 0.526 & 0.497 & 0.585 & 0.580 & 0.524 & 0.537 \\
 & Time (s) $\downarrow$ & 3.16 & \textbf{4.47} & 7.38 & 5.05 & 5.91 & 22.27 & 8.48 & 8.19 & 18.96 & \textbf{27.71} & 12.44 & 8.11 \\
\midrule
\multirow{4}{*}{BC} & Completion (\%) $\uparrow$ & 96.7 & 97.6 & 93.6 & 90.2 & 92.5 & 85.7 & 82.4 & 91.5 & 87.2 & 91.6 & 86.9 & 91.3 \\
 & Precision (\%) $\uparrow$ & 93.4 & 93.6 & 81.1 & 74.2 & 80.8 & 71.4 & 60.6 & 77.0 & 70.2 & 79.8 & 70.3 & 79.0 \\
 & Force MAE (N) $\downarrow$ & 0.590 & 0.763 & 0.707 & 0.866 & 0.801 & 0.943 & 0.753 & 0.764 & 0.792 & 0.605 & 0.856 & 0.773 \\
 & Time (s) $\downarrow$ & \textbf{3.10} & 4.54 & 7.26 & 5.26 & 6.01 & 22.31 & 8.59 & 8.45 & 18.84 & 28.48 & 12.52 & 8.19 \\
\midrule
\multirow{4}{*}{RL} & Completion (\%) $\uparrow$ & 97.5 & \textbf{99.7} & \textbf{97.1} & \textbf{95.0} & \textbf{95.8} & \textbf{89.8} & \textbf{86.8} & \textbf{96.4} & \textbf{91.6} & \textbf{92.9} & \textbf{91.6} & \textbf{94.8} \\
 & Precision (\%) $\uparrow$ & \textbf{96.6} & \textbf{99.0} & \textbf{90.6} & \textbf{86.5} & \textbf{89.3} & \textbf{80.3} & \textbf{69.6} & \textbf{89.3} & \textbf{80.9} & \textbf{89.4} & \textbf{83.0} & \textbf{87.6} \\
 & Force MAE (N) $\downarrow$ & \textbf{0.310} & \textbf{0.325} & \textbf{0.369} & \textbf{0.348} & \textbf{0.380} & \textbf{0.438} & \textbf{0.396} & \textbf{0.331} & \textbf{0.433} & \textbf{0.454} & \textbf{0.402} & \textbf{0.362} \\
 & Time (s) $\downarrow$ & 3.13 & 4.73 & \textbf{7.09} & \textbf{4.66} & \textbf{5.65} & \textbf{20.58} & \textbf{7.72} & \textbf{7.78} & \textbf{17.48} & 27.84 & \textbf{11.28} & \textbf{7.67} \\
\bottomrule
\end{tabular*}
\caption{Task-level and zero-shot results with 50 matched seeds per task. Completion, precision, force MAE, and time are reported on separate rows. The mean covers the eight primary tasks; the three unseen columns are zero-shot evaluations. Bold marks the best controller for each task and metric.}
\label{tab:summary}
\end{table}

Table~\ref{tab:summary} shows that RL raises mean completion to 94.8\%, from 91.3\% for BC and 92.0\% for IK. BC itself is below IK in both completion and precision, confirming its role as an initialization rather than the final controller. Relative to BC, RL raises precision from 79.0\% to 87.6\%, reduces force MAE by 53\% from 0.773 to 0.362\,N, and reduces time from 8.19 to 7.67\,s. The IK and BC traces exhibit visible stroke jitter, which leaves ink outside the target and depresses precision; outcome RL suppresses this effect (Appendix Fig.~\ref{fig:case_study_app}). RL improves completion on seven of eight primary tasks and precision and force error on all eight. Its gains on unseen English words, Chinese idioms, and Chinese characters show execution of new requested trajectories rather than memorization. Appendix~\ref{app:extended_results} provides complete metrics and diagnostics.

\section{Conclusion}
We present a force-aware grasp-to-writing system. A transferred grasp trunk retains the pen, all-task IK-guided BC initializes writing, and outcome RL refines the controller. RL improves completion, precision, force, and time over BC and IK, including on unseen English and Chinese content.
\label{main:end}

\clearpage
\bibliography{refs}
\clearpage
\appendix
\section{Implementation and evaluation details}
\label{app:details}
\paragraph{Reading the supplementary material.}
This appendix provides implementation details, task definitions, training settings, ablations, and extended results for the reported trajectory-conditioned system.

\begin{figure}[h]
\centering
\includegraphics[width=\textwidth]{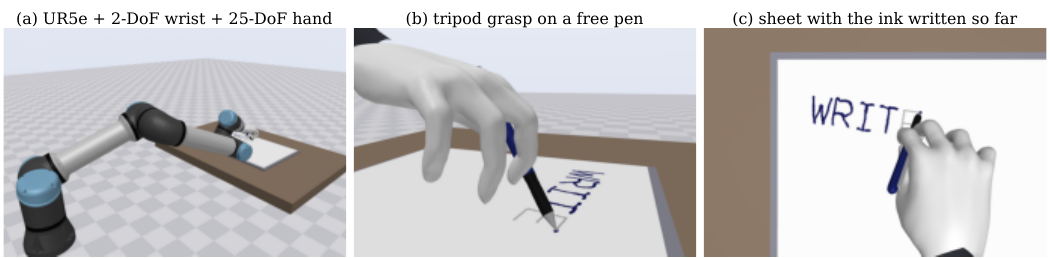}
\caption{Simulated setup. (a) A UR5e arm with a 2-DoF wrist and a 25-DoF anthropomorphic hand writes on paper with a free pen. (b) The tripod grasp. (c) Top view mid-episode; gray marks the remaining target.}
\label{fig:env_app}
\end{figure}

Figure~\ref{fig:env_app} shows the simulated scene. Hand and environment settings are listed in Table~\ref{tab:environment}; network, training, and evaluation settings are listed separately in Table~\ref{tab:settings}. The physical control contract is therefore distinguished from choices made by the learning algorithm.

\begin{table}[h]
\centering\small
\begin{tabularx}{\textwidth}{lX}
\toprule
Parameter & Hand and environment configuration \\
\midrule
Hand and arm & High-DoF anthropomorphic robotic hand mounted on a UR5e assembly \\
Actuated coordinates & Seven arm coordinates, two wrist joints, and 25 finger joints \\
Physical action interface & Seven integrated arm-position increments; 27 speed--torque-limited wrist/finger duties \\
Control frequency & 25\,Hz \\
Tool and initial grasp & Free rigid-body pen, initially seated in a tripod grasp; retained through contact \\
Grip assistance & No virtual grip assistance in the reported writing configuration \\
Starting height & 2--6\,mm in the reported writing configuration \\
Trajectory advancement & Tip-driven pointer; pen-down proximity tolerance 1.5\,mm; skip timeout 0.25\,s \\
Task set & Line, circle, shape, digits, letters, word, cursive, and Chinese writing \\
Content split & Shared training/evaluation content pool; primary numerical comparison uses the eight-task suite \\
\bottomrule
\end{tabularx}
\caption{Hand, environment, and task-interface settings for the reported writing study. These specify the physical control problem; model architecture and optimization settings are given separately in Table~\ref{tab:settings}.}
\vspace{-0.5cm}
\label{tab:environment}
\end{table}

\clearpage
\begin{table}[h]
\centering\small
\begin{tabularx}{\textwidth}{lX}
\toprule
Parameter & Model, training, and evaluation configuration \\
\midrule
\multicolumn{2}{l}{\textbf{Model architecture and observation processing}} \\
Input representation & 137 state entries plus 16 trajectory samples of five entries each; samples every two trajectory indices; no image encoder \\
Actor & Separate grasp and writing MLPs, each $217\to256\to256$ with tanh activations; masked linear heads \\
Policy output & 34 physical-action entries plus one learned stop output; stop threshold 0.5 \\
Critic & Independent $217\to256\to256\to1$ MLP \\
Normalization & Vector statistics frozen during final RL; normalized observations clipped to $[-10,10]$ \\
\midrule
\multicolumn{2}{l}{\textbf{Behavior cloning}} \\
Demonstration noise & Clean and noisy demonstrations, $\sigma=0.01,0.02$ \\
BC objective & Arm/wrist loss with inverse-variance weighting \\
BC optimizer & Adam, learning rate $3\times10^{-4}$, batch 256, 60 epochs per candidate \\
\midrule
\multicolumn{2}{l}{\textbf{Reinforcement learning}} \\
Grasp preservation & Grasp-mean parameters frozen; Gaussian log-standard-deviations remain trainable \\
Rollout collection & 512 environments $\times$ 1024 control steps; asynchronous execution \\
Actor update & PPO clip 0.2; full-rollout gradient accumulation; up to four passes; approximate KL target 0.03 on 16,384 randomly sampled transitions \\
Optimizer and returns & Adam $10^{-5}$ with linear decay; minibatch 2048; $\gamma=1$, GAE $\lambda=0.95$; reward scale 0.001 \\
Critic update & Three initial critic-only iterations, followed by two critic passes per rollout \\
Initial exploration & Log-standard-deviations: arm/wrist $-4.6$, fingers $-3.9$, stop $-3.0$; floor $-6.0$ \\
Recorded budget & 39,845,888 transitions; 193.8 minutes for the reported final RL run \\
\midrule
\multicolumn{2}{l}{\textbf{Evaluation protocol}} \\
Policy execution & Deterministic learned mean and the network's own stop output \\
Evaluation sample & Primary comparison: 50 seeds $\times$ eight tasks \\
\bottomrule
\end{tabularx}
\caption{Model and learning settings, separated from the hand/environment parameters in Table~\ref{tab:environment}. BC settings describe the available collection/training recipe, not a reconstructed per-shard manifest of the final clone. Evaluation settings specify measurement rather than training updates.}
\label{tab:settings}
\end{table}

\paragraph{Data and checkpoint provenance.}
The main comparison uses an eight-task evaluation with 50 episodes with different seeds per task for IK, all-task BC, and final RL. Intermediate evaluations in the learning plot are snapshots of the same final RL run.

\paragraph{Evaluation-suite changes.}
The development code merged the original content pools, and new evaluation-only word, idiom, and Chinese-character pools were constructed for the zero-shot tests. A later script-font update and the folding of cursive into the word task change the distribution relative to the primary eight-task evaluation. 

\clearpage
\subsection{Training and checkpoint details}
\label{app:curriculum}
The reported handwriting method transfers a learned grasp, collects IK writing demonstrations over the complete repertoire, fits one BC policy on the pooled data, and refines the composed policy with full-task outcome RL. Figure~\ref{fig:pipeline} shows this separation.

\begin{table}[h]
\centering\small
\begin{tabularx}{\textwidth}{lXX}
\toprule
Component & Training condition & Training and transfer \\
\midrule
Grasp controller & 240 teacher episodes for BC, then pen-hold PPO under motion and contact & Transfer the selected learned grasp into the MLP \\
Writing data and BC & Traces, glyphs, words/cursive, and Chinese together & IK arm/wrist plus learned grasp supplies clean/noisy demonstrations; fit one all-task clone \\
Composed-policy RL & Full writing mixture with grasp and contact dynamics active & Both branches execute; refine outcomes while retaining grasp-mean weights \\
\bottomrule
\end{tabularx}
\caption{Transferred grip controller, all-task writing demonstration collection and BC, and full-task composed-policy RL.}
\end{table}

\paragraph{Recorded grasp lineage.}
The reported handwriting checkpoint transfers an existing grasp policy initialized from 240 teacher episodes and followed by 8M PPO steps over static hold, moving hold, and environment-driven scribble. The later trajectory-based implementation folds the blank-map encoder contribution and constant input entries into MLP biases.

\begin{figure}[h]
\centering
\includegraphics[width=\textwidth]{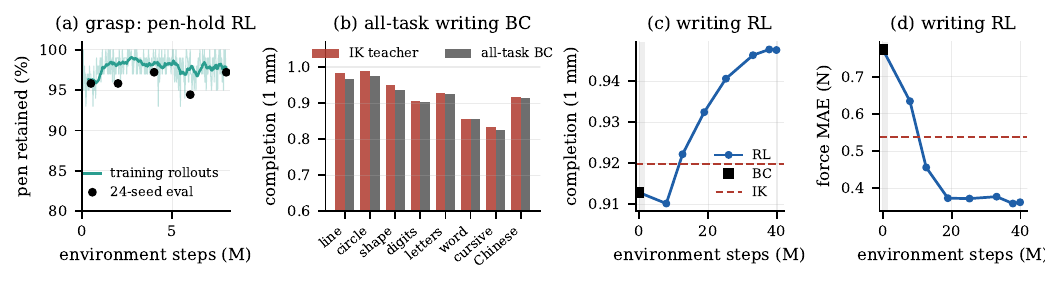}
\caption{Training across the pipeline. (a) Pen-hold RL from the BC grasp: fraction of training episodes that keep the pen (thin: per iteration; thick: 15-iteration mean) and 24-seed evaluations over static hold, moving hold, and scribble. (b) Per-task completion of all-task BC and its IK teacher (50 seeds). (c, d) Writing RL completion and force MAE over environment steps; BC at step 0, IK dashed.}
\label{fig:pipeline_curves_app}
\end{figure}

\paragraph{All-task writing BC.}
The clone is fit once on the complete pool of traces, glyphs, words, cursive, and Chinese characters, and initializes final RL. IK supplies demonstrations but does not act during learned-policy evaluation. Figure~\ref{fig:pipeline_curves_app} shows training progress for each stage.

\paragraph{Outcome-based RL for writing.}
BC provides a broad initialization, but matching IK commands does not directly optimize the written trace or contact force. We therefore refine the writing branch with PPO while executing the full grasp--pen--arm system.

\clearpage
\subsection{Detailed policy, imitation, and RL implementation}
\label{app:implementation}
\paragraph{Trajectory-conditioned control.}
The input comprises 137 retained state entries (proprioception, tactile feedback, task and geometric state) and 16 future trajectory samples. Each sample contains tip-relative displacement, relative height, requested force, and nominal time offset. Samples are spaced by two trajectory indices. A tip-driven pointer advances through the request with a 1.5\,mm proximity tolerance and a 0.25\,s skip timeout; the request therefore does not advance solely according to wall-clock time. Pen-down input waypoints specify the sheet surface, rather than exposing IK's below-surface pressing offset. The policy thus receives explicit task geometry.

\paragraph{Separate grasp and writing features.}
Let $x_t\in\mathbb{R}^{217}$ be the normalized input and $m$ select the arm, wrist, and stop dimensions. The Gaussian mean is
\begin{equation}
\mu(x_t)=(1-m)\odot g_{\theta_g}(x_t)+m\odot w_{\theta_w}(x_t).
\end{equation}
The grasp branch inherits the learned holding controller; the writing branch is initialized through imitation. The selected outputs form 34 physical controls and one stop entry, requesting termination above 0.5. A separate $217\to256\to256\to1$ critic has the same hidden activations. The implementation contains 383,082 parameters including inactive head rows and Gaussian log-standard-deviations. During RL, grasp-mean parameters and observation statistics remain fixed; log-standard-deviations remain trainable, including those for fingers. 

\paragraph{Imitation initialization.}
IK supplies arm/wrist commands while a learned holding policy supplies finger commands. The collection recipe gathers clean rollouts and two noise-injected sets ($\sigma=0.01,0.02$ in executed action units), retaining clean IK labels. We use DART-style noise injection~\cite{laskey2017dart} with fixed variance. The delivered arm/wrist clone trains on the pooled demonstrations from all tasks and uses a weighted squared-error loss,
\begin{equation}
\mathcal L_{\mathrm{BC}}=\mathbb E_{(x,a^*)}\sum_{d=1}^{9}
 w_d\bigl(\mu_d(x)-a_d^*/s_d\bigr)^2,\quad
 w_d=\operatorname{clip}\!\left[\left(\frac{\operatorname{median}_j\sigma_j}
 {\max(\sigma_d,10^{-3})}\right)^2,1,100\right],
\end{equation}
where $s_d$ is the action scale and $\sigma_d$ the demonstration-label standard deviation. The recorded recipe uses Adam, learning rate $3\times10^{-4}$, batch 256, and 60 epochs per candidate. The writing branch is merged with the grasp branch and the stop output is fitted on rollouts. The resulting all-task clone is the BC baseline below.

\paragraph{Rewarding drawing outcomes.}
The final reward configuration uses
\begin{equation}
 r_t=\lambda_C\Delta C_t-\lambda_P\Delta P_t-\lambda_E\Delta E_t
       -\lambda_\tau\Delta\tau_t+\lambda_q\bigl(\|p_{t-1}-q_t\|-\|p_t-q_t\|\bigr)-D_t.
\label{eq:reward}
\end{equation}
\begin{table}[h]
\centering
\caption{Weights in the final drawing-outcome reward.}
\label{tab:reward_weights}
\small
\begin{tabular}{ccl}
\toprule
Weight & Value & Reward term \\
\midrule
$\lambda_C$ & 1000 & Ordered target coverage \\
$\lambda_P$ & 500 & Off-target ink \\
$\lambda_E$ & 200 & Contact-force MAE \\
$\lambda_\tau$ & 300 & Normalized elapsed time \\
$\lambda_q$ & 5000 & Progress toward the guidance point \\
\bottomrule
\end{tabular}
\end{table}
Here $C_t$ is online ordered target coverage within 1\,mm; $P_t$ accumulates off-target ink normalized by expected pen-down samples, capped at 2; $E_t$ is running contact-force MAE; and $\Delta\tau_t$ is elapsed time divided by nominal task duration. The guidance point $q_t$ is the first uncovered glyph point at or after the coverage pointer, placed 1.5\,mm below the sheet. The same $q_t$ is used on both sides of the distance difference.  $D_t$ accounts for revoking accumulated coverage reward on a drop and charging remaining time for specified premature terminations. Coverage is an online surrogate for final alignment; RL has no IK-action or timing loss.

\paragraph{Stable policy refinement.}
We optimize the PPO clipped objective~\cite{schulman2017ppo} using gradients accumulated over each full $512\times1024$ rollout. Up to four actor passes use Adam at $10^{-5}$ with linear decay. Proposed parameter displacements are rescaled if approximate KL on 16,384 randomly sampled rollout transitions exceeds 0.03; this is an empirical budget, not a hard KL guarantee. The critic receives two minibatch passes (batch 2048), after three initial critic-only iterations. We use $\gamma=1$, GAE $\lambda=0.95$, clip 0.2, reward scale 0.001, and initial arm/wrist noise $\exp(-4.6)\approx0.01$. The recorded run collects 39.85M transitions in 193.8 minutes.

\clearpage
\section{Task catalogue, force-annotated strokes, and the 5D cursor}
\label{app:task_config}
This section describes the task interface used by the reported system. It distinguishes the three holding tasks used for grasp learning from the seven writing tasks in the current suite and the three evaluation-only generalization tasks. The primary comparison used eight writing tasks, with cursive separate from word. Task geometry, target force, and trajectory timing jointly define the requested behavior. The current source catalogue is not a claim that archived checkpoints were retrained on every subsequent task change.

\subsection{Tasks used in grasp and writing training}
\begin{figure}[h]
\centering
\includegraphics[width=\textwidth]{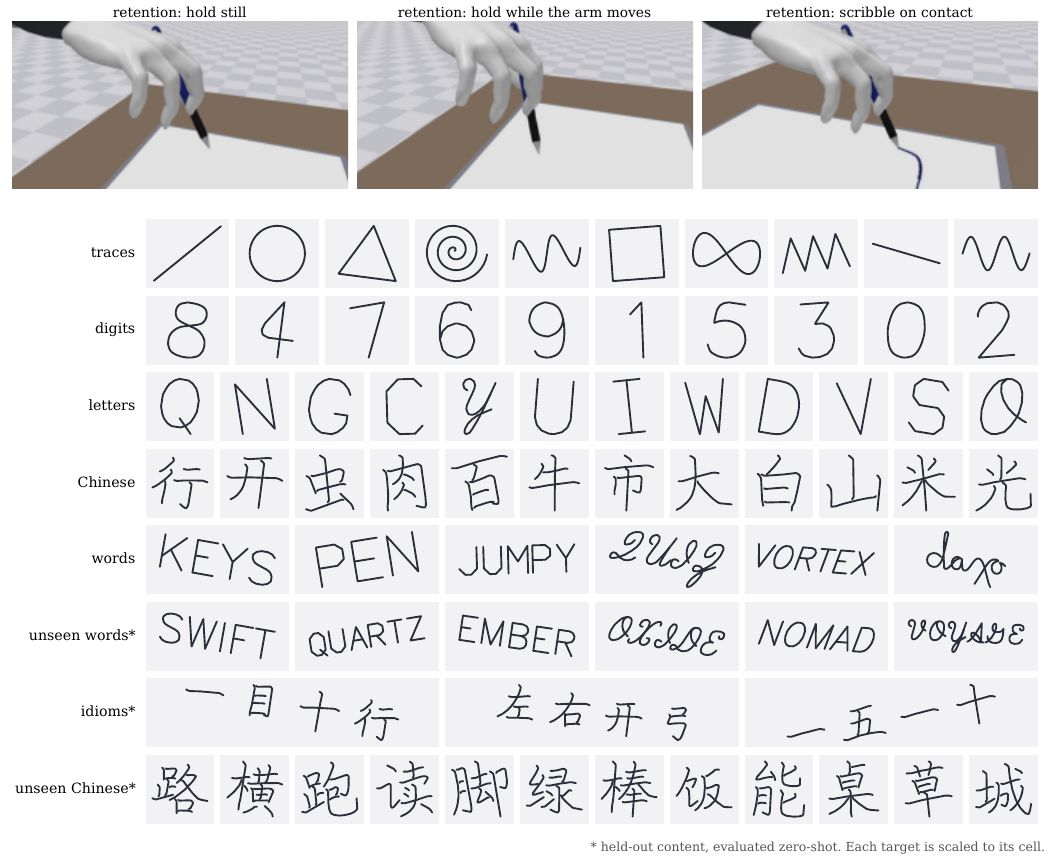}
\caption{Task suite. Top: the three pen-retention tasks. Bottom: one target per tracing and writing task, drawn to scale with its size; * marks held-out content evaluated zero-shot.}
\label{fig:task_suite_app}
\end{figure}

\begin{table}[h]
\centering\small
\begin{tabularx}{\textwidth}{lp{1.2cm}X}
\toprule
Task ID & Nominal duration & Requested behavior and configuration \\
\midrule
\multicolumn{3}{l}{\textbf{L0: grasp training; arm/wrist motion supplied by the environment}} \\
\texttt{L0\_hold\_static} & 6\,s & Retain the seated pen while the hand is still; no target ink strokes. \\
\texttt{L0\_hold\_moving} & 8\,s & Retain the pen during imposed transport and wrist disturbance: 11 waypoints, motion scale 2.2, arm-motion box half-extents $(0.12,0.08,0.085)$\,m, wrist shake amplitude $4^\circ$. \\
\texttt{L0\_scribble} & 10\,s & Retain the pen during an environment-driven paper-contact path: 14 waypoints, 3\,mm commanded press depth, $8^\circ$ wrist wobble, and 15\,N impact limit. This is a contact-retention task, not glyph matching. \\
\midrule
\multicolumn{3}{l}{\textbf{L1: geometric tracing}} \\
\texttt{L1\_trace\_line} & 8\,s & A 60--120\,mm straight line with randomized orientation and placement; flat force profile. \\
\texttt{L1\_trace\_circle} & 10\,s & A circle 40--90\,mm in diameter; flat force profile. \\
\texttt{L1\_trace\_shape} & 12\,s & A sampled shape, 40--90\,mm requested height: line, circle, square, triangle, spiral, sine, figure-eight, or zigzag; flat or ramp force profile. \\
\midrule
\multicolumn{3}{l}{\textbf{L2: handwriting}} \\
\texttt{L2\_write\_digits} & 10\,s & One digit, 50--65\,mm requested height, with a recorded-pressure profile. Script-selected digits use the sans face in the current suite. \\
\texttt{L2\_write\_letters} & 10\,s & One capital letter, 50--65\,mm requested height, sampled font, and recorded-pressure profile. \\
\texttt{L2\_write\_word} & $(8+3n)$\,s & An $n$-character word from the content pool; requested height 20--35\,mm, sampled font, slant 0 or 0.3, and recorded-pressure profile. \\
\texttt{L2\_write\_chinese} & 12\,s & One Chinese character from the at-most-six-stroke pool, requested height 50--65\,mm, and recorded-pressure profile. \\
\midrule
\multicolumn{3}{l}{\textbf{E2: evaluation-only content, excluded from the training task list}} \\
\texttt{E2\_write\_word\_unseen} & $(8+3n)$\,s & One of 20 held-out English words, with the word-task geometry and force profile. \\
\texttt{E2\_write\_idiom} & 48\,s & One of ten four-character Chinese idioms composed of trained characters; character height 50--65\,mm before width fitting. \\
\texttt{E2\_write\_chinese\_unseen} & 12\,s & One of 122 characters outside the trained pool; every character has at least seven strokes, extending both content and stroke complexity. \\
\bottomrule
\end{tabularx}
\caption{Task catalogue relevant to this paper, from the current samplers. Dimensions are requested geometric ranges before width fitting. Nominal duration paces the trajectory; it is not necessarily the final elapsed episode time. All listed L1/L2/E2 writing samplers use the pen configuration.}
\label{tab:task_catalogue}
\end{table}
Figure~\ref{fig:task_suite_app} shows the retention tasks and one target per tracing and writing task.

\paragraph{Placement and variation.}
Stroke templates are normalized, scaled, optionally sheared, rotated, and translated on the sheet. The common placement routine samples rotation in $[-0.18,0.18]$ radians unless overridden, respects a 30\,mm margin, and reduces scale when needed to fit the width. Content, font, geometry, and force variation depend on the particular task; they should not all be assumed independent for every sampler. The current font pool contains block, sans, rounded sans, and script. Current content pools merge earlier training/test pools.

\paragraph{Nominal time versus episode horizon.}
The \texttt{duration} field sets the nominal trajectory budget. For writing with \texttt{end\_mode=policy}, the implementation permits up to twice this nominal duration, with earlier termination possible through the policy's stop output, dropping, or applicable termination rules. Thus a word's $(8+3n)$\,s configuration should not be interpreted as its measured writing time or as an unconditional hard timeout.

\paragraph{Additional registered tasks.}
The broader environment also registers finger-only scribbling and pen pickup at L0; brush strokes, Chinese brush writing, and brush layouts at L3; sketch hatching at L4; shape/rainbow/scene painting at L5; and chopstick holding, cylinder, cube, block, and sphere tasks at L6. These are outside the reported writing comparisons. The pen-writing idiom result uses the dedicated evaluation-only E2 sampler rather than the unrelated brush-layout task.

\clearpage
\subsection{What a task configuration specifies}
\label{app:task_schema}
A task is sampled as a \texttt{TaskSpec}, rather than specified only by its natural-language instruction. Its ordered strokes, force annotations, metadata, and timing determine the actual control request.

\begin{table}[h]
\centering\small
\begin{tabularx}{\textwidth}{lp{2.2cm}X}
\toprule
Field & Type / unit & Meaning \\
\midrule
\texttt{task\_id}, \texttt{level} & Identifier & Select the task family and level. \\
\texttt{instruction} & Text & Human-readable description. The policy MLP does not encode this string as language. \\
\texttt{strokes} & List of $N_j\times2$ arrays, m & Ordered centerline points in paper coordinates relative to the sheet center. List boundaries distinguish pen-down strokes. \\
\texttt{forces} & List of $N_j$ arrays, N & Requested pressing force at each corresponding stroke point. \\
\texttt{widths} & List of $N_j$ arrays, mm & Target mark widths derived from the force arrays and the tool's ink model. \\
\texttt{duration} & Seconds & Nominal schedule duration used to pace the requested strokes and transits. \\
\texttt{tool}, \texttt{start} & Enumerations & Tool identity and seated-grasp versus table-start initialization. The reported writing tasks use a seated pen. \\
\texttt{hold\_motion} & Mode or null & Selects externally driven static/moving/scribble behavior for holding tasks. \\
\texttt{hold\_scale}, \texttt{hold\_waypoints} & Scale / count & Set imposed-motion amplitude and number of waypoints in holding tasks. \\
\texttt{hold\_box}, \texttt{hold\_jitter\_deg} & m / degrees & Arm transport bounds and wrist-disturbance amplitude. \\
\texttt{scribble\_depth\_m}, \texttt{scribble\_wrist\_deg} & m / degrees & Commanded contact depth and wrist wobble for driven scribble. \\
\texttt{impact\_limit\_N} & N & Contact-impact limit used by the relevant task's success criterion. \\
\texttt{meta} & Dictionary & Sampled text, font, slant, shape, or other task-specific metadata. \\
\bottomrule
\end{tabularx}
\caption{Task-definition fields. Reward weights, optimizer settings, and network architecture are separate from this schema. Stored legacy success thresholds should not be confused with the paper's reported 1\,mm completion metric.}
\label{tab:task_schema}
\end{table}

\paragraph{Force-annotated target strokes.}
For stroke $j$, write its configuration as
\begin{equation}
\begin{aligned}
 S_j&=\{(x_{j,i},y_{j,i})\}_{i=1}^{N_j}, &
 F_j&=\{f_{j,i}\}_{i=1}^{N_j},\\
 W_j&=\{w_{j,i}\}_{i=1}^{N_j}, &
 w_{j,i}&=\operatorname{inkWidth}(\texttt{tool},f_{j,i}).
\end{aligned}
\end{equation}
The force arrays are target commands, not measurements of the force the robot actually achieves. Their correspondence with stroke points makes force part of the requested writing content, alongside path and order. Width is derived from the same force values, rather than sampled as an unrelated visual decoration.

The default pen force-profile generator samples a base force between 0.9 and 2.5\,N. A \emph{flat} profile holds one sampled force along a stroke. A \emph{ramp} modulates it monotonically along normalized arc length. A \emph{recorded-pressure} profile samples a stored Wacom pressure curve, interpolates it along the stroke's arc length, and rescales it into 0.9--2.5\,N. The general generator also supports a \emph{swell} profile; it is not used by the listed L1/L2 samplers. All generated values are clipped to $[0.15,4.0]$\,N. In particular, the base-force range is not a strict bound on the ramp or swell profile before clipping.

\paragraph{From force to the target image and observed ink.}
The environment can rasterize a task map using
\begin{equation}
 I^{\star}(x,y)\approx255\,\frac{f^{\star}(x,y)}{F_{\max}},\qquad F_{\max}=4\,\mathrm{N},
\end{equation}
with integer clipping to $[1,255]$ on target pixels and zero background. Stroke thickness follows \texttt{widths}. The stored intensity is a force encoding; display software may invert it to render larger values as darker ink. Actual ink instead depends on simulated contact force and tip speed, so target intensity is not a guarantee of the deposited appearance. The current deposition model makes force affect width and darkness, with speed-dependent adjustments. Although these maps are useful for visualization and evaluation, the \texttt{traj2} actor consumes the numerical trajectory window, not map pixels.

\clearpage
\subsection{The 5D cursor and its lookahead window}
\label{app:cursor5d}
\paragraph{Absolute task-space request.}
The environment expands the ordered 2D strokes into a time-indexed 3D schedule containing approach, descent, pen-down motion, lifted transitions, and a final pen-up phase. Each requested sample can be represented as a five-dimensional cursor
\begin{equation}
 c_k^{\star}=(x_k^{\star},y_k^{\star},h_k^{\star},f_k^{\star},t_k^{\star}).
\end{equation}
Here $x,y$ are horizontal target coordinates, $h$ is height above the sheet, $f$ is requested force, and $t$ is nominal due time. ``5D'' therefore means position, force, and time; it is not a five-joint action or a five-DoF end-effector pose. The source stores these components across timeline/height/force arrays and constructs the five-entry representation for the observation.

For a pen-down request, exposed height is zero and force is positive. For a pen-up/transit request, force is zero and the requested height follows the lift trajectory. The teacher's 1.5\,mm below-surface pressing target is not exposed as a negative target height: the task specifies surface contact and desired force separately. At overlapping strokes, the current implementation obtains timeline force through the spatial target-force lookup; it does not supply an additional stroke-specific force identifier to the actor.

\begin{table}[h]
\centering\small
\begin{tabularx}{\textwidth}{lllX}
\toprule
Entry & Absolute request & Policy entry & Interpretation \\
\midrule
1 & $x_k^{\star}$ & $x_k^{\star}-p_{t,x}$ & Horizontal displacement, m \\
2 & $y_k^{\star}$ & $y_k^{\star}-p_{t,y}$ & Horizontal displacement, m \\
3 & $h_k^{\star}$ & $h_k^{\star}-h_t$ & Height error relative to the sheet, m \\
4 & $f_k^{\star}$ & $f_k^{\star}$ & Requested force, N; not force error or measured contact force \\
5 & $t_k^{\star}$ & $t_k^{\star}-t$ & Due-time offset, s; negative when the nominal due time has passed \\
\bottomrule
\end{tabularx}
\caption{Semantics of one five-dimensional trajectory/cursor sample. Normalization is applied after assembling the policy vector.}
\label{tab:cursor5d}
\end{table}

\paragraph{What the policy actually sees.}
The actor observes a window, not just one current cursor. Let $k_t$ be the current task pointer and $K$ the last valid timeline index. With 16 points and stride 2,
\begin{align}
 k_{t,i}&=\min(k_t+2i,K),\qquad i=0,\ldots,15,\\
 z_{t,i}&=(x_{k_{t,i}}^{\star}-p_{t,x},\ y_{k_{t,i}}^{\star}-p_{t,y},\
 h_{k_{t,i}}^{\star}-h_t,\ f_{k_{t,i}}^{\star},\ t_{k_{t,i}}^{\star}-t).
\end{align}
The resulting $16\times5=80$ values are flattened and appended to the 137 retained state entries, yielding 217 inputs. At 25\,Hz, adjacent samples are 0.08\,s apart in nominal schedule time and the last is 1.2\,s beyond the first. This is schedule spacing, not a guaranteed 1.2-second horizon relative to the current physical execution, because the pointer can lag. Near the end, indices are clamped to the last point. For holding tasks without a writing timeline, the trajectory window is zero.

\paragraph{Pointer advancement versus teacher timing.}
The reported configuration uses \texttt{adv\_mode=tip}. A pen-down timeline point passes the advancement check when planar tip error is at most 1.5\,mm and tip height is at most 0.8\,mm above the sheet. The height condition is a geometric contact proxy, not a requirement that measured normal force exceed a threshold. Pen-up points use a 6\,mm planar tolerance and a 15\,mm vertical-error tolerance. If no point advances, the current point is skipped after the 0.25\,s wait threshold. Skipping does not itself deposit ink or earn completion credit. Consequently, pointer progress, actual stroke coverage, and measured force remain distinct quantities.



\clearpage
\section{Extended results and diagnostics}
\label{app:extended_results}

\subsection{Metric definitions and task-level effects}
Ordered completion uses monotone alignment between the requested stroke sequence and deposited ink, with a 1\,mm threshold. Precision is the fraction of ink samples within 1\,mm of the target. Symmetric Chamfer distance measures geometric proximity, while the reported matched-stroke discrete Fr\'echet distance permits reversal when pairing and averages matched pairs. Force MAE compares eligible contact-force samples with the requested force. Elapsed time includes learned stopping or failure termination, and drop rate is the percentage of episodes ending in a detected pen drop.

\subsection{Case study: RL suppressing stroke jitter}
\begin{figure}[h]
\centering
\begin{tikzpicture}
\node[inner sep=0] (case) {\includegraphics[width=\textwidth]{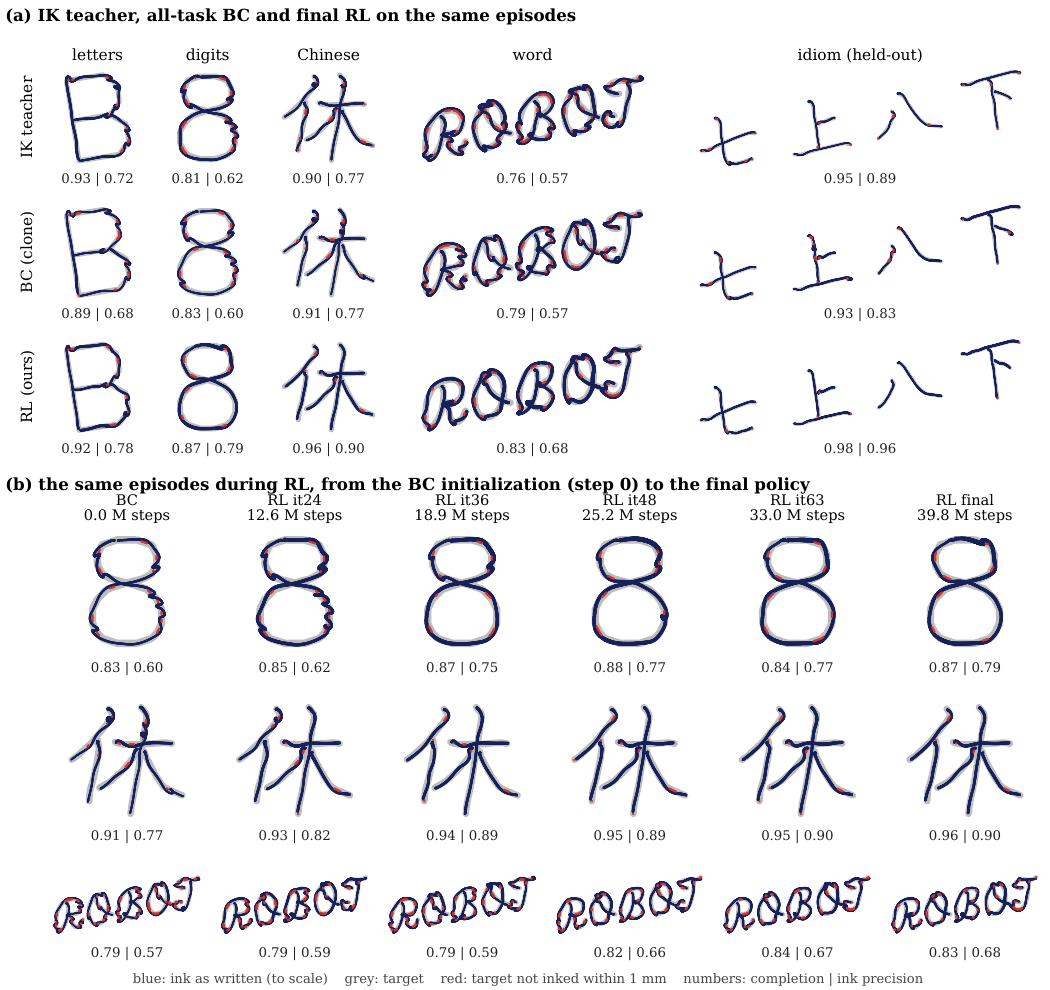}};
\end{tikzpicture}
\caption{Matched episodes across IK, BC, and final RL (top), and their evolution from BC initialization through RL (bottom). Blue is deposited ink, gray is the target, and red marks target points not covered within 1\,mm. Numbers report completion $\mid$ ink precision. The high-frequency deviations in IK and BC leave off-target ink; RL produces smoother, more target-aligned strokes.}
\label{fig:case_study_app}
\end{figure}

The gains extend across the full repertoire. Digits improve from 90.6\% IK completion to 95.0\%, Chinese from 91.7\% to 96.4\%, and words from 85.4\% to 89.8\%. The line is the exception: its completion is 97.5\% for RL and 98.2\% for IK. Seed 26254 produces no ink for both BC and RL after an initial tip excursion of roughly 50\,mm and a hover termination. Empty ink receives a 50\,mm Chamfer cap, increasing its influence on the aggregate. We retain this episode in all reported means.

\subsection{Progress during final RL}
\begin{figure}[h]
\centering
\includegraphics[width=\textwidth]{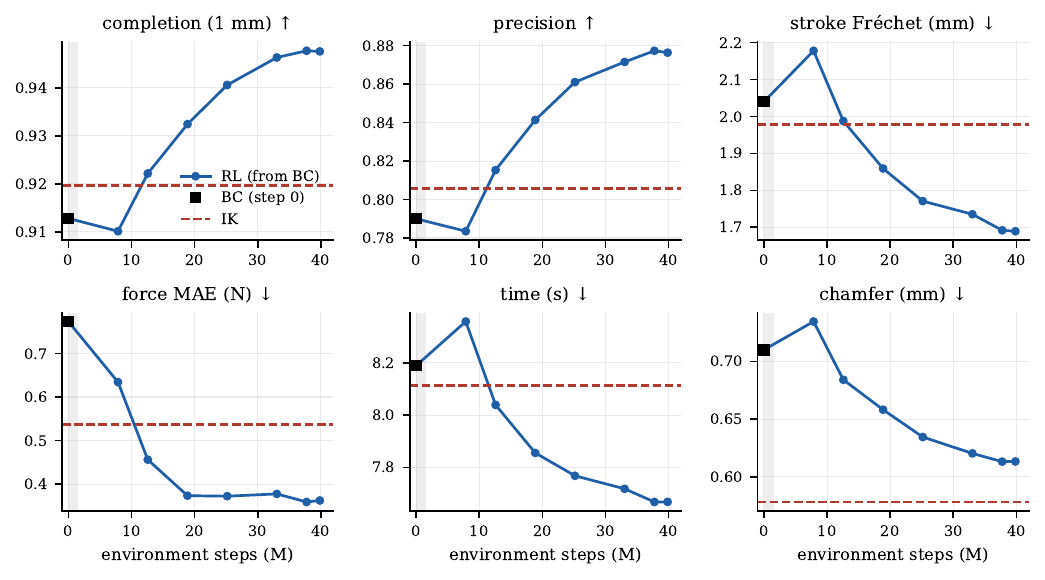}
\caption{Fifty-seed evaluations during final RL. The initial point is BC and dashed lines indicate IK on the same seeds. Snapshot timing makes the horizontal axis approximate by at most one rollout. All checkpoints come from one training run.}
\label{fig:learning_app}
\end{figure}

The transition from imitation to RL is nonmonotone: early evaluations can lose completion or shape quality before later checkpoints improve completion, ink precision, force control, and elapsed time together. This motivates evaluating physical outcomes throughout refinement rather than treating a rising return as sufficient evidence of better writing.

\begin{figure}[h]
\centering
\includegraphics[width=\textwidth]{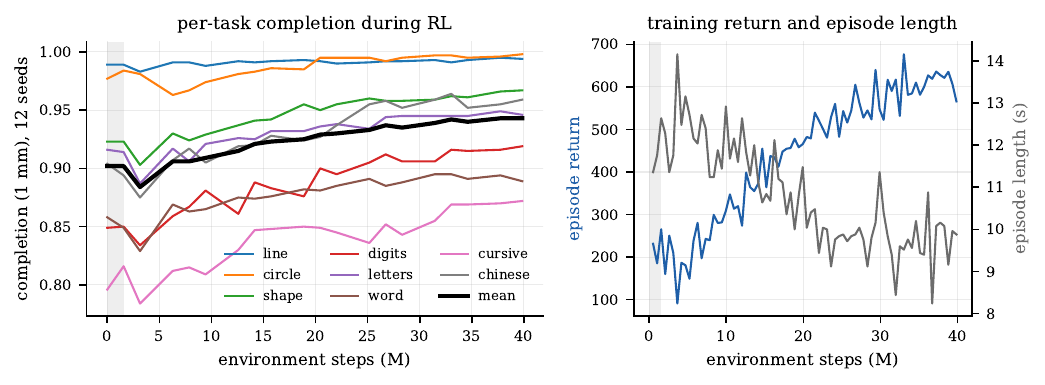}
\caption{Twelve-seed task monitoring and training return/episode length during final RL. This monitoring set differs from the 50-seed aggregate evaluation and is descriptive rather than a repeated-training statistic.}
\label{fig:task_curves_app}
\end{figure}

The 12-seed BC monitoring mean is 90.2\%, whereas the 50-seed BC mean is 91.3\%. These baselines are not interchanged when quantifying improvement. Intermediate checkpoints are snapshots of one final RL run, not independent training seeds.




\end{document}